\documentclass{article}

\usepackage{graphicx}
\usepackage[table]{xcolor}

\usepackage{booktabs}

\usepackage[preprint]{corl_2021} 

\title{Autonomy 2.0: \\ Why is self-driving always 5 years away?}

\author{
  Ashesh Jain, Luca Del Pero, Hugo Grimmett, Peter Ondruska\\
  Lyft Level 5 self-driving\\
  \texttt{\{asheshjain, ldelpero, hgrimmett, pondruska\}@lyft.com} \\
}

\begin{document}
\maketitle


\begin{abstract}
Despite the numerous successes of machine learning over the past decade (image recognition, decision-making, NLP, image synthesis), self-driving technology has not yet followed the same trend. In this paper, we study the history, composition, and development bottlenecks of the modern self-driving stack. We argue that the slow progress is caused by approaches that require too much hand-engineering, an over-reliance on road testing, and high fleet deployment costs. We observe that the classical stack has several bottlenecks that preclude the necessary scale needed to capture the long tail of rare events.  
To resolve these problems, we outline the principles of \emph{Autonomy 2.0}, an ML-first approach to self-driving, as a viable alternative to the currently adopted state-of-the-art. This approach is based on (i) a fully differentiable AV stack trainable from human demonstrations, (ii) closed-loop data-driven reactive simulation, and (iii) large-scale, low-cost data collections as critical solutions towards scalability issues. We outline the general architecture, survey promising works in this direction and propose key challenges to be addressed by the community in the future.

\end{abstract}



\begin{figure}[h]
   \centering
   \includegraphics[width=\textwidth]{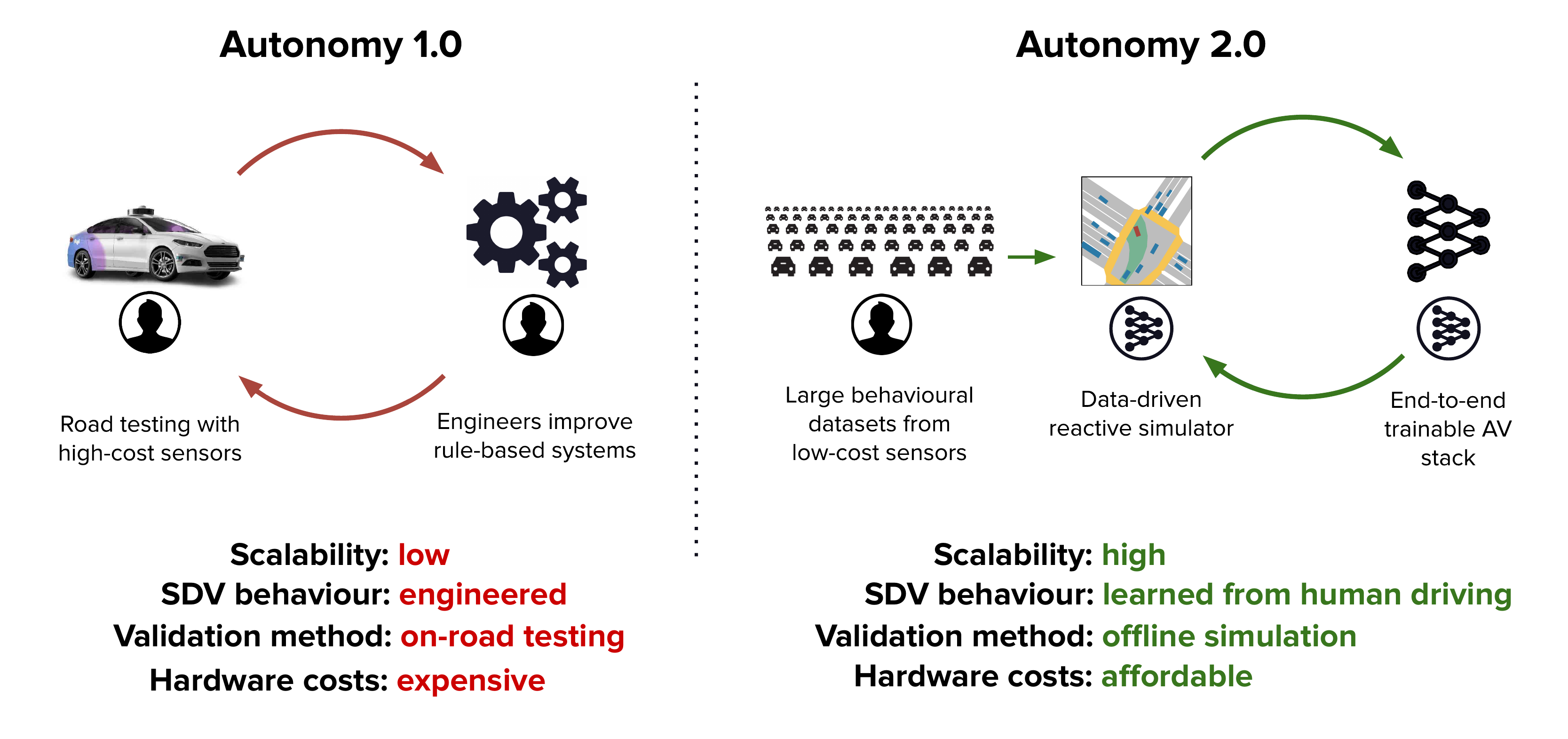}
   \caption{The development process of a typical state-of-the-art autonomy stack (Autonomy 1.0) vs proposed ML-first stack (Autonomy 2.0). We argue that the current self-driving industry progress is slow due to an inefficient human-in the loop development. These issues can be solved by training a differentiable self-driving stack in a closed-loop simulation constructed out of a large collection of human driving demonstrations. }
   \label{fig:a1vsa2}
\end{figure}

\section{Introduction}
\label{sec:intro}
Self-Driving Vehicles (SDVs) have been an active research area for decades and make regular headlines since the DARPA Grand Challenges in 2005-2007. Many companies are attempting to develop the first level 4+ SDVs \cite{autonomy_levels}, some for more than a decade. We have seen small-scale SDV testing, but despite the numerous unrealised predictions that ubiquitous SDVs are `only 5 years away' \cite{2012_prediction, 2015_prediction, 2016_prediction}, production-level deployment still seems a distant future. Given the limited rate of progress, some questions naturally arise: (a) why did the research community underestimate the problem's difficulty? (b) are there fundamental limitations in today's approaches to SDV development?

\begin{figure}[b]
   \centering
   \includegraphics[width=0.9\textwidth]{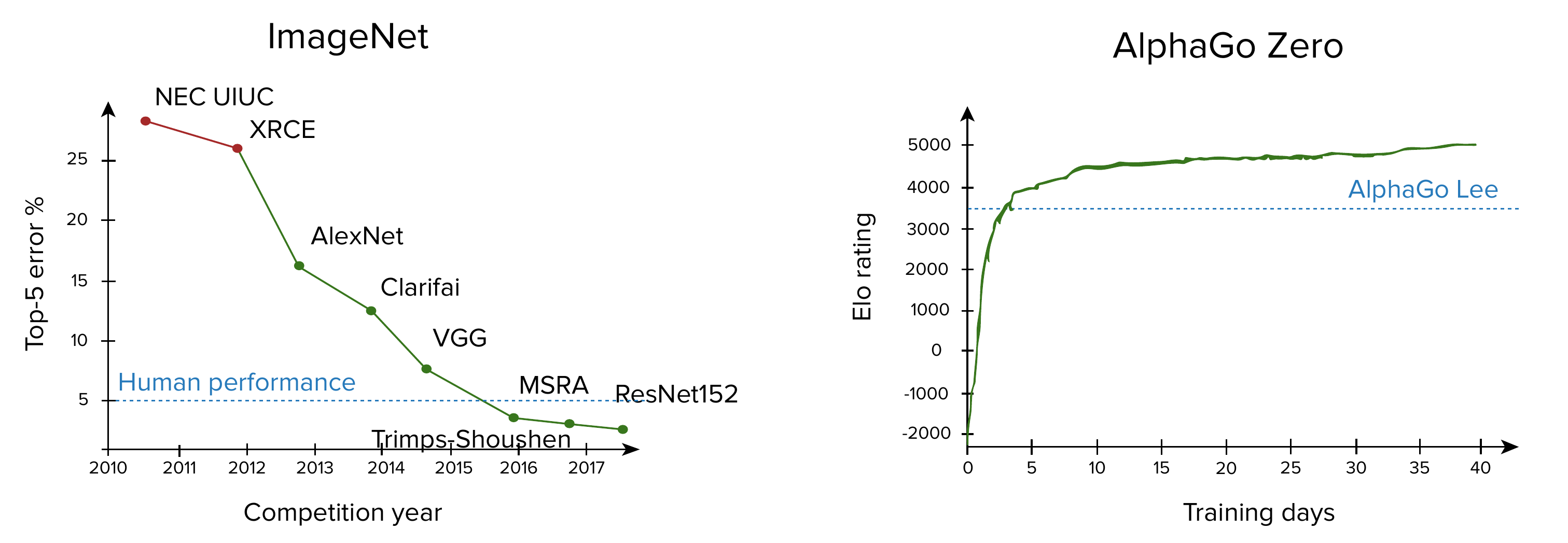}
   \caption{Two successes of ML: ImageNet [left] and AlphaGo Zero [right] achieving super-human performance in perception and decision making. Self-driving has not followed this trend yet. We argue this is because Autonomy 1.0 self-driving technology adopted across the industry is not fully data enabled causing scalability bottlenecks.}
   \label{fig:success}
\end{figure}

After the DARPA challenges, most of the industry decomposed the SDV technology stack into HD mapping, localisation, perception, prediction, and planning \cite{10.5555/1121596}. Following breakthroughs enabled by ImageNet \cite{deng2009imagenet}, the perception and prediction parts started to become primarily machine-learned. However, behaviour planning and simulation are still largely \emph{rule-based}: performance improves using humans writing increasingly detailed rules that dictate how the SDV should drive. There has been the belief that, given very accurate perception, rule-based approaches to planning may suffice for human-level performance. We refer to this approach as \emph{Autonomy 1.0}.

Production-level performance requires significant \emph{scale} to discover and appropriately handle the 'long tail' of rare events. We argue that Autonomy 1.0 will not achieve this due to three scalability bottlenecks: (i) rule-based planners and simulators do not effectively model the complexity and diversity of driving behaviours and need to be re-tuned for different geographical regions, they fundamentally have not benefited from the breakthroughs in deep-learning (Fig.~\ref{fig:success}) (ii) due to the limited efficacy of rule-based simulators, evaluation is done mainly via road-testing, which lengthens the development cycle, and (iii) road testing operations are expensive and scale poorly with SDV performance.

Our proposed solution to these scale bottlenecks is to turn the entire SDV stack into an ML system that can be trained and validated offline using a large dataset of diverse, real-world data of human driving. We call this \emph{Autonomy 2.0}. Autonomy 2.0 is a data-first paradigm: ML turns all parts of the stack (including planning and simulation) into data problems, and performance improves with better data sets rather than by designing new driving rules (Fig.~\ref{fig:a1vsa2}). This unlocks the scalability required for mastering the long tail of rare events and scaling to new geographies. All that is needed is to collect large enough datasets and retrain the system.

The key challenges to Autonomy 2.0 are (i) formulating the stack as an end-to-end differentiable network, (ii) validating it offline in a closed-loop with a machine-learned simulator, and (iii) collecting the large amounts of human driving data required to train them.

In the next section, we explain and analyse the bottlenecks in Autonomy 1.0 as typically adopted in the industry, then explain the benefits of Autonomy 2.0, and end by highlighting the open questions that must be answered for SDVs to \emph{truly} be 5 years away.


\section{Autonomy 1.0}
\label{sec:sota}

\begin{figure}[t]
   \centering
   \includegraphics[width=\textwidth]{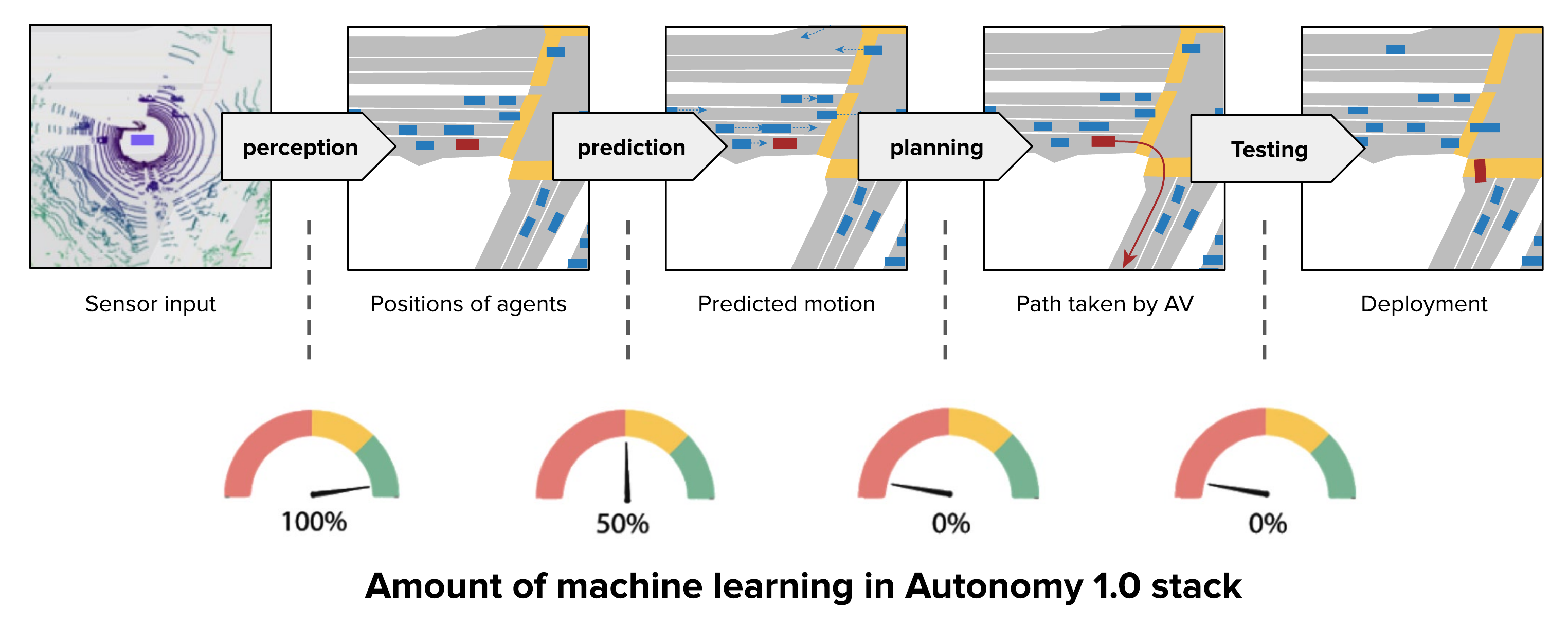}
   \caption{Autonomy 1.0 -- the typical technology stack, indicating the amount of ML used in individual components. While perception and prediction are ML-based, planning and simulation still rely on non-scalable rule-based systems (sec.~\ref{sec:bottlenecks})}
   \label{fig:cycle}
\end{figure}

In this section we survey the state-of-the-art Autonomy 1.0 technology used in a typical self-driving program. The typical stack (Fig.~\ref{fig:cycle}) is composed of these components: perception, prediction and planning, which subsequently answer the questions of \textit{what is around the car?} \textit{what is likely to happen next?} and \textit{what should the car do?}. An essential part of the development cycle of the stack is testing, which answers \textit{"what is the performance of the system?"}

\textbf{Perception}. The typical solution consists of neural networks processing raw camera \cite{ssd-6d}, LIDAR \cite{qi2016pointnet, voxelnet} and radar data \cite{meyer2019-radar}. Observations from different sensors are implicitly or explicitly fused \cite{kocic2018sensors, cho2014multi}, producing the 3D positions and other attributes (e.g. size, orientation) of the \emph{traffic participants} around the SDV (e.g. vehicles, pedestrians and cyclists). These systems are trained on hand-labelled sensor data \cite{lyft2019}. They also typically rely on HD maps \cite{nuscenes2019, lyft2019} containing information about the static environment and its traffic rules.

\textbf{Prediction}. Today, most rule-based methods that extrapolate the observed motion of traffic participants \cite{social-forces-98} are being replaced with ML motion forecasting, where neural network are trained to predict a future trajectory from a few seconds of observations \cite{Lee2017DESIREDF, sophiegan, multipath, sriram2021}. These predictions are used to estimate future space occupancy and the likelihood of collisions.

\textbf{Planning}. Despite a lot of progress in the field of reinforcement learning \cite{dqn, alphago}, Autonomy 1.0 still heavily relies on \emph{rule-based} trajectory optimisation techniques. These techniques combine global search and local optimisation to find a trajectory that minimises an expert-designed cost function~\cite{Ziegler08, Jr.00rrt-connect:an, Bandyopadhyay13, Montemerlo2009}, which is then fed to a controller converting it into speed and steering signals. The cost function consists of various comfort and safety terms hand-engineered by domain experts to achieve a desirable behaviour, such as average acceleration, distance to other vehicles, adherence to traffic rules, etc. 

\textbf{Testing} estimates \textbf{\textit{``what is the performance of the system?''}} Typically, the system performance is road-tested by deploying it under the supervision of a safety driver, who is ready to disengage the autonomy system at any time, i.e. take control in the case of incorrect behaviour. Such disengagements are an important part of the development cycle (Fig. ~\ref{fig:a1vsa2}). They provide a valuable source of information on which part of the self-driving stack to improve. For example, errors caused by planning lead to modifying the expert cost function by adding new weights.

\section{Scalability bottlenecks of Autonomy 1.0}
\label{sec:bottlenecks}

The Autonomy 1.0 stack described in the previous section can perform well under regular conditions \cite{darpa, Montemerlo2009}. Attaining L4-L5 production-level performance, however, requires scaling the paradigm to cover the 'long tail' of rare events such as road closures, road accidents, other agents breaking the rules-of-the-road etc. At the same time, the solution needs to scale to multiple cities with diverse agent behaviours. In this section we explore the scalability bottlenecks that make it challenging for Autonomy 1.0 to handle this long tail of events.

\subsection{Trying to capture complex and diverse behaviours with rule-based systems}
Rule-based systems seek to optimise a cost function designed by domain experts who write new terms and balances their weights. New scenarios require new terms and weights, and so development requires large engineering teams and is a major bottleneck. Operating SDVs in a new location can mean person-years of re-tuning. Designing interfaces between the Autonomy 1.0 components also come with a high engineering and performance cost. Even though they provide human-interpretable abstraction between different tasks, they limit the ability to express the nuance of the encountered driving situations.


\subsection{Reliance on road-testing and low-realism offline simulation}
On-road testing provides the highest possible realism, and safety driver disengagements can give actionable feedback to improve the system. However, the time between cost-engineering and road testing makes for a long development cycle: days or weeks. Moreover, it is non-reproducible: one cannot repeat scenarios exactly, preventing direct comparison between versions.

One strategy for offline evaluation is to replay sensor logs to evaluate the performance of a new system version. However, if the SDV actions during replay are different to those in the log, other traffic participants do not react accordingly. E.g. an SDV, driving slower than in the original log can cause a false positive rear-end collision. This hinders fair comparison between software versions.

A degree of reactivity can be achieved by adding rule-based behaviours to the other traffic participants or employing a synthetic simulator \cite{carla}. However, rule-based simulation has the same limitations as rule-based planning: hand encoding diverse and realistic behaviours for pedestrians, cyclists, and other vehicles scales poorly. 

\subsection{Limited fleet deployment scale}
As requirements on system performance, scenario rarity, and the statistical significance of evaluation increase, so does the amount of required road-testing. Increasing fleet size can speed up the development cycle. Still, Autonomy 1.0 SDVs rely on expensive sensors and compute to satisfy the stringent requirements on perception accuracy imposed by rule-based planning. This makes SDVs expensive, and so the available budget puts an upper bound on the development speed and statistical performance guarantees.

\section{Autonomy 2.0}
\label{sec:autonomy2}



In this section, we propose Autonomy 2.0, an ML-first approach to self-driving focused on achieving high scalability. It is based on three key principles: i) closed-loop simulation, which is learned from the collected real-world driving logs; ii) decomposing SDV into an end-to-end differentiable neural network; and iii) the data needed to train the planner and simulator is collected at a large-scale using commodity sensors.




\subsection{Data-driven closed-loop reactive simulations}

\begin{figure}[t]
   \centering
   \includegraphics[width=\linewidth]{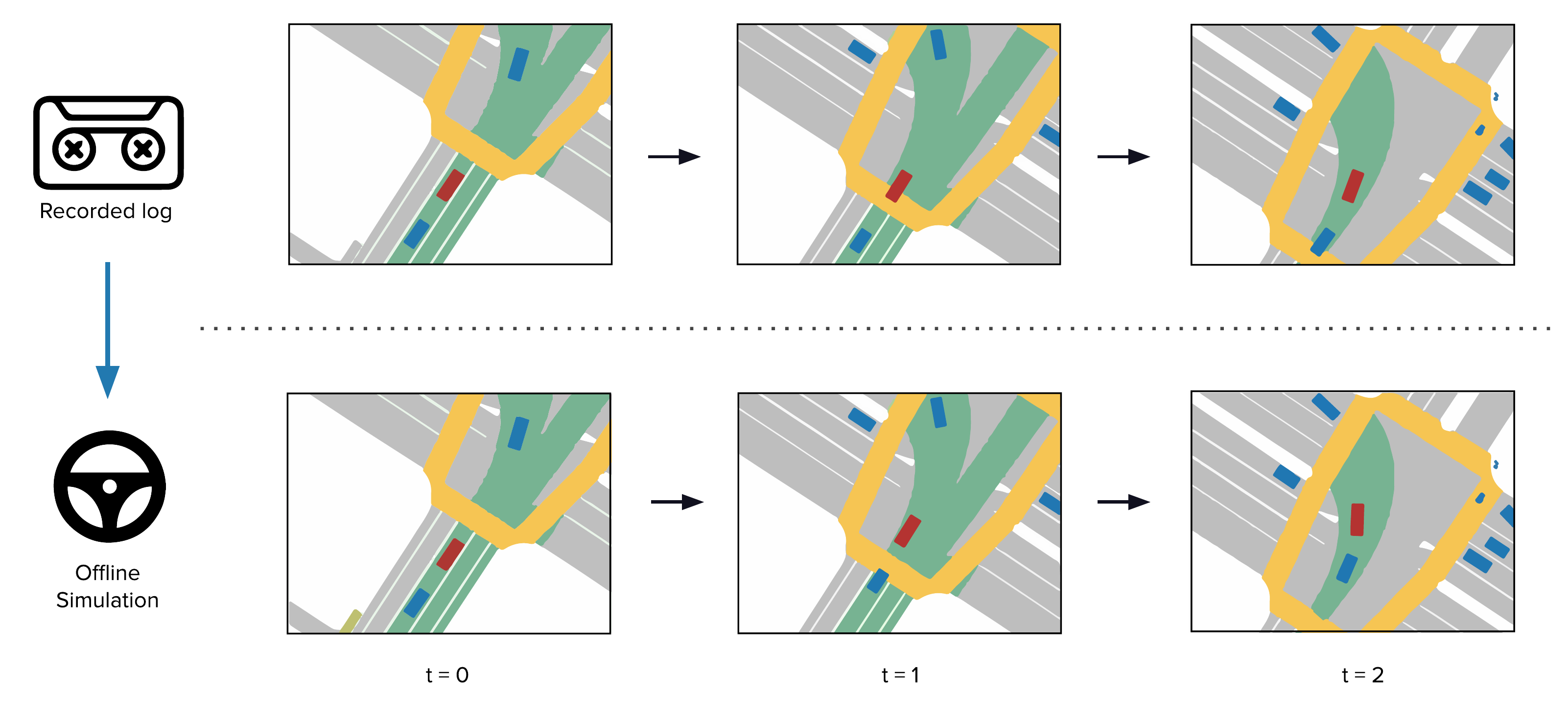}
   \caption{A data-driven reactive simulator \cite{bergamini2021simnet} enables the synthesis of new, realistic driving scenarios based on a recorded log. This allows the training and evaluation self-driving stack offline, without the need for extensive road testing (sec.~\ref{sec:sim})}
   \label{fig:simulator}
\end{figure}

\label{sec:sim}
Most of the evaluation in Autonomy 2.0 is done offline in simulation. This is in contrast to the Autonomy 1.0 reliance on road testing due to the limitations of rule-based simulation. It does not mean ceasing road-testing altogether, but its purpose is less prominent in the development cycle and is mainly used to verify the simulator performance. For simulation to be an effective replacement of road testing for development, it needs three properties: 
\begin{enumerate}
   \item an appropriate simulation state representation for the task;
   \item the ability to synthesise diverse and realistic driving scenarios with high fidelity and reactivity;
   \item when applied to new scenarios and geographies, performance increases with the amount of data.
\end{enumerate}


Simulation outcomes must be very realistic since any difference between simulation and reality would result in inaccurate performance estimates, but it does not need to be photo-realistic \cite{Dosovitskiy17} and instead focus only on the representation of the planner. We reason that that in order to achieve a high level of realism, the simulation itself must be learned directly from the real world. Recently, \cite{bergamini2021simnet} showed how realistic and reactive simulations can be constructed from previously collected real-world logs using birds-eye-view representations. As shown in Fig.~\ref{fig:simulator}, this simulation can be then deployed to turn any log into a reactive simulator for testing autonomous driving policies. The appealing property of such an approach is that simulation fidelity scales with data. By collecting more logs, the coverage of simulation proportionally grows as each log can be turned into a simulation case. 
However, further work needs to be done to address behavioural multi-modality since we cannot know the intentions of other traffic participants ahead of time, and the same action from an SDV can lead to many different outcomes. 
There are initial works \cite{rhinehart2021, multipath, trafficsim}, but more work is required (Sec.~\ref{sec:openqs}).

\subsection{A fully differentiable stack trained from human demonstrations}
\label{sec:differentiable_stack}

Autonomy 1.0 has rule-based components that are hand-engineered, as are the human-interpretable interfaces between perception, prediction, planning, and simulation (Sec.~\ref{sec:sota}). Unlike Autonomy 1.0, the Autonomy 2.0 stack is fully trainable from human demonstrations, and so its complexity scales in proportion to the amount of training data. 
In order to train such a system, several conditions need to be met:
\begin{enumerate}
   \item each component, including planning, needs to be trainable and end-to-end differentiable;
   \item trainable using human demonstrations;
   \item performance scales with the amount of training data.
\end{enumerate}

End-to-end differentiability is essential as it allows to back-propagate gradients across the stack. Perception and prediction are already differentiable in Autonomy 1.0, but planning is not. Recently, \cite{Bansal_2018_RSS} have shown that a high-capacity driving policy can be represented by a neural network and trained from a collection of human demonstrations. The appealing property is this avoids the need for hand-engineering of cost function since the training signal is implicitly given by the demonstrations.

\begin{figure}
   \centering
   \includegraphics[width=\textwidth]{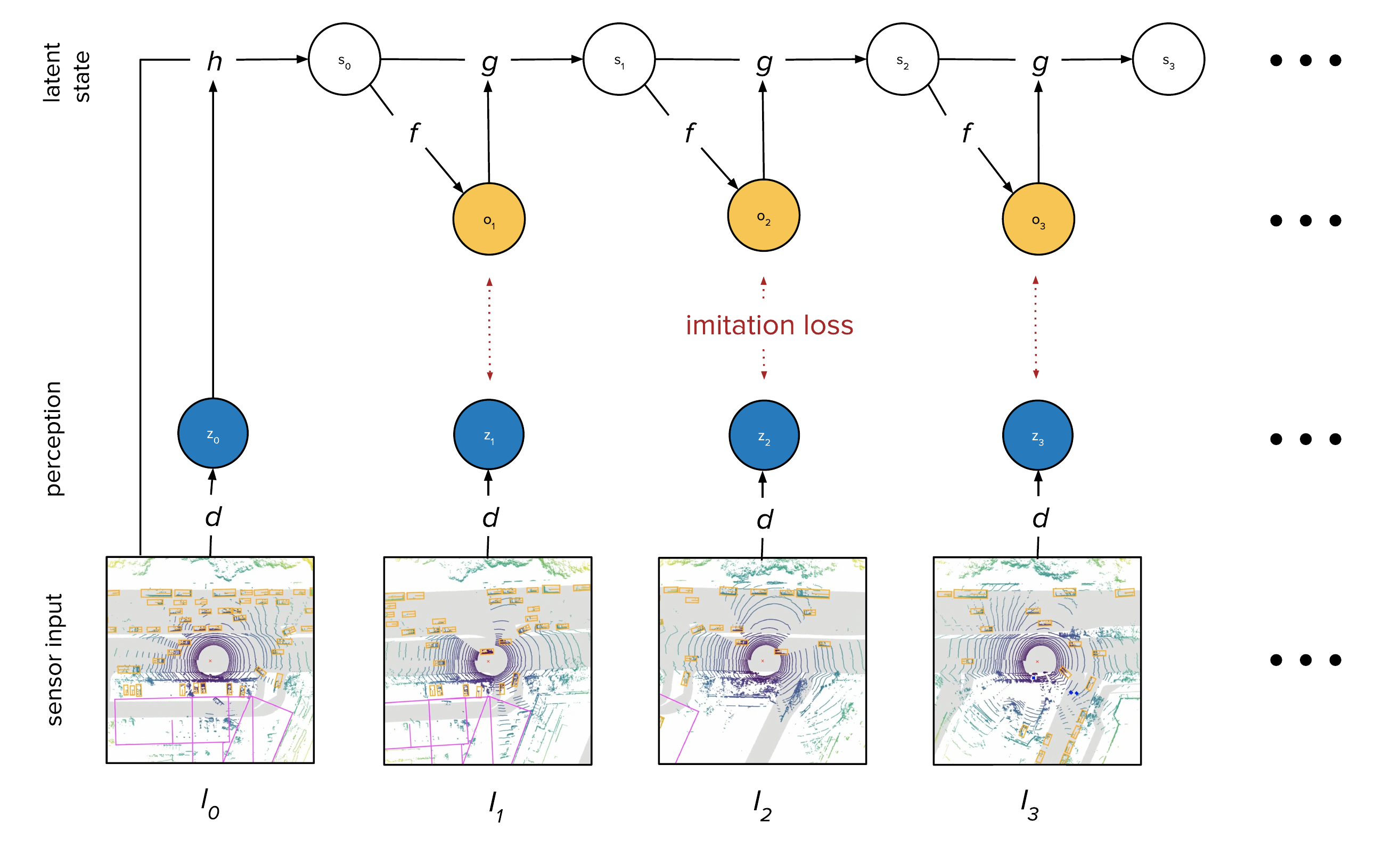}
   \caption{Architecture of a fully differentiable Autonomy 2.0 stack motivated by \cite{alphazero} (sec.~\ref{sec:differentiable_stack}) that can be trained end-to-end from data without the need of engineering individual blocks and interfaces.
   $d$, $h$, $f$, and $g$ are learnable neural networks. $d$ and $h$ give a latent representation of the scene on which planning happens. $f$ represents the policy of the SDV and the agents in the scene. $g$ is the state transition function function. $I_0$ is the input to the network, while $\{I_1,\cdot \cdot,I_3\}$ provides supervision during training.}
   \label{fig:unroll}
\end{figure}

Training the system requires proper formulation based on the nature of sequential decision-making tasks. Imitation learning, while simple, suffers from covariate shift caused by the accumulation of errors \cite{ross2011reduction}. 
To implicitly avoid the problem of covariate shift and causal confusion~\cite{causal-levine} we argue that the system needs to be trained in a closed loop with the reactive data-driven simulation outlined Sec.~\ref{sec:sim}. We can generate new driving episodes under a current policy without further data collection and then update the policy based on the imitation loss \cite{GAIL}. 

Furthermore, inspired by \cite{alphazero} we propose that the networks for each component (i.e. driving policy, perception, and simulation) can be coupled and trained together as outlined in Fig.~\ref{fig:unroll}. The policy $f$ generates the action that the SDV should take at a given state $s$, and a (learnable) simulator $g$ (Sec.~\ref{sec:sim}) transitions the state conditioned on the action and the current state. One way to enable co-training is for the entire stack to be fully differentiable. 
The ML planner is then plugged into a differentiable architecture and is trained jointly with perception. Importantly, it allows the quality of perception output to be determined in a way that is most suitable for training driving policies, rather than setting a more arbitrary perception requirement of centimetre-level accurate 3D bounding boxes. The interpretability property, though, is still maintained as the learned perception representation can be back-projected into interpretable space for examination. This approach to formulating the autonomy problem has many similarities to MuZero \cite{muzero}.

\subsection{Large-scale low-cost data collection}

The systems discussed so far use human demonstrations as training data, i.e. sensor data with the corresponding trajectory chosen by a human driver serving as supervision. To unlock production-level performance, these data need to have:
\begin{enumerate}
   \item enough size and diversity to include the long tail of rare events (Sec.~\ref{sec:bottlenecks});
   \item sufficient sensor fidelity, i.e. the sensors used to collect the data need to be accurate enough to train the planner and simulator effectively and
  \item cheap enough to collect at this scale and fidelity.
\end{enumerate}

While recently we saw the first public datasets with human demonstrations, these are limited to a few thousand miles of data~\cite{houston_2020_corl}. Observing the long tail will likely require collecting hundreds of millions of miles of data because most driving is uneventful, e.g. there are roughly 5 crashes per million miles driven in the US~\cite{BTS}. 

Collecting these volumes constrains us to rely on crowd-sourcing, as using cars and drivers dedicated solely for data collection is not economically viable at this scale. It also constrains us to use only low-cost commodity sensors like cameras. These, however, produce lower perception accuracy than high fidelity sensors (like HD LIDAR) used in Autonomy 1.0 to power rule-based planning. Hence, in choosing sensors for data collection, we face a trade-off between the scalability (how much data we collect) and the fidelity (the resulting perception accuracy) we can expect from them.

\begin{table}[t]
   \centering
   \def\arraystretch{1.3}
   \begin{tabular}{@{}l c l@{}}
     \hline\toprule
     \textbf{Sensor type} & \textbf{Perception accuracy on KITTI} & \textbf{Scalability}\\ 
      & \textbf{(car mAP, moderate)} & \\ 
     \hline 
     Monocular cameras & 22$\%$~\cite{Luo_2021_CVPR} & High \\
     Stereo cameras & 52$\%$~\cite{Chen_2020_CVPR} & Medium\\
     Stereo + sparse LIDAR & 64$\%$~\cite{Wang_2019_CVPR}& Medium\\
     HD LIDAR & 75$\%$~\cite{Lang_2019_CVPR} & Low \\
     \bottomrule\hline
   \end{tabular}
   \vspace{3mm}
   \caption{When collecting data, we face a trade-off between sensor scalability and fidelity, which impacts directly perception accuracy. Scalable sensors like cameras (top two rows) achieve lower accuracy than the expensive ones traditionally used in SDVs (e.g. LIDAR, bottom row). Recent progress in perception algorithms has reduced this gap, and solutions with cameras and sparse LIDAR (third row) have become really competitive on standard perception benchmarks like KITTI~\cite{Geiger_2013_IJRR}}
   \label{tab:lfd}
\end{table}



Which sensors should we use? Recent progress in perception algorithms showed a reduced gap in perception accuracy between HD and commodity sensors like cameras ~\cite{Luo_2021_CVPR,Chen_2020_CVPR} and sparse LIDAR~\cite{Wang_2019_CVPR} on the KITTI benchmark~\cite{Geiger_2013_IJRR} (Table~\ref{tab:lfd}). Moreover, the actual perception accuracy requirements for training Autonomy 2.0 systems~\cite{philion_2020_CVPR} might be relaxed by training perception and planning jointly (Sec.~\ref{sec:differentiable_stack}). This is corroborated by recent results suggesting that training ML planners on large amounts of data with lower accuracy perception might be preferable to smaller amounts with higher accuracy~\cite{Chen_2021_ICRA}. As the sensor prices stand today, this points to the camera only (and optionally sparse LIDAR) as the most promising solution to strike a balance between cost and fidelity, but this remains an open problem. 

\section{Open questions}
\label{sec:openqs}
In this work, we outlined the Autonomy 2.0 paradigm, which is designed to solve self-driving using an ML-first approach. By removing the human-in-the-loop, this paradigm is significantly more scalable, which we argue is the main limitation for achieving high SDV performance. While promising, this direction leaves many open questions that are yet to be answered, such as:
\begin{itemize}
\item What is the appropriate state representation for simulation and planning? How much should it be learned vs interpretable?
\item How should we measure the probability of scenarios? How should we detect outlier, never-before-seen cases? 
\item What are the limits of what can be trained offline from human demonstrations vs need real-time reasoning using search?
\item How much do we need to simulate? How should we measure the performance of the offline simulation itself?
\item How much data do we need to train high-performing planning and simulation components? What sensors should we use for large-scale data collection?
\end{itemize}
Answering these questions is critical to self-driving and other real-world robotic problems and can stimulate the research community to unlock high-performing SDVs sooner rather than later.

\label{sec:conclusion}



\clearpage
\acknowledgments{We would like to thank following contributors who work on Autonomy 2.0 at Lyft Level 5: Moritz Niendorf, Qiangui Huang, Oliver Scheel, Matt Vitelli,
Błażej Osiński, Ana Ferreira, Luca Bergamini, Ray Gao, Yawei Ye, Sammy Sidhu, Lukas Platinsky, Christian Pereone, Jasper Friedrichs, Maciej Wołczyk, Long Chen, and Sacha Arnoud.}


\bibliography{example}  
\end{document}